\documentclass[10pt]{article}
\usepackage[margin=0.82in]{geometry}
\usepackage{amsmath,amssymb,mathtools}
\usepackage{booktabs}
\usepackage{graphicx}
\usepackage{microtype}
\usepackage{natbib}
\usepackage{hyperref}
\usepackage{enumitem}
\usepackage{array}
\usepackage{url}
\usepackage[T1]{fontenc}
\usepackage{lmodern}

\graphicspath{{figures/}}
\hypersetup{colorlinks=true,linkcolor=black,citecolor=black,urlcolor=black}
\setlist{nosep,leftmargin=*}
\newcommand{\E}{\mathbb{E}}
\newcommand{\R}{\mathbb{R}}
\newcommand{\norm}[1]{\left\lVert #1\right\rVert}

\title{Hidden Boundary Motion in Transformer Optimization:\\
Function-Space Orthogonalization of Affine Weight and Bias Updates}
\author{Zhang Gongyue, Sheng Yixuan,Liu donghan, Wang Zhiyong, Ren Weihong and Liu honghai}

\date{}

\begin{document}
\maketitle

\begin{abstract}
Weights and biases are normally optimized as separate parameter tensors, yet they do not represent separate functions when the input to an affine layer has nonzero mean. For an affine map $z=Wx+b$ with input mean $\mu$, a weight update contains a sample-independent displacement $\Delta W\mu$ that is functionally indistinguishable from a bias update. We call this hidden contribution \emph{boundary motion} and decompose each update into a centered, sample-varying \emph{shape} component and a shared \emph{boundary} component. On a four-layer Transformer trained from scratch on IMDb, the bias-like term $g_b\mu^\top$ has a median norm equal to 0.664 of the raw weight-gradient norm across affine layers and training checkpoints. More strikingly, the median ratio $\norm{\Delta W\mu}/\norm{\Delta b}$ is 134.7, while $\norm{\Delta W\mu}/\norm{\Delta b+\Delta W\mu}$ is 0.994. Thus, under AdamW, the observed boundary motion is almost entirely realized through the weight matrix rather than the explicit bias.

We implement a diagnostic optimizer, Shape--Boundary Orthogonal AdamW (SBO-AdamW), that optimizes $g_W-g_b\mu^\top$ and $g_b$ with independent Adam states and compensates the weight-induced boundary displacement. In a single-seed experiment, SBO-AdamW raises validation accuracy from 81.68\% to 85.81\% and validation-selected test accuracy from 78.73\% to 82.73\%, with the best validation checkpoint occurring at step 800 instead of step 3000. However, the moving-batch-center compensation produces severe bias-coordinate drift and strongly reduces boundary energy. The present evidence therefore supports hidden boundary motion as an important optimization mechanism, but it does not yet establish a final general-purpose optimizer. A stable centered-affine parameterization is identified as the required next step.
\end{abstract}

\section{Introduction}
Modern optimization methods assign different treatments to different parameter tensors. Adam and AdamW apply coordinate-wise second-moment normalization \citep{kingma2014adam,loshchilov2017adamw}; matrix-aware methods such as Muon instead exploit the geometry internal to two-dimensional weight matrices \citep{liu2025muon}. Biases are generally left as ordinary vectors. This division is computationally convenient, but it assumes that parameter type coincides with functional role.

For an affine layer,
\begin{equation}
    z = Wx+b,
\end{equation}
the assumption is false whenever the layer input has nonzero mean. Let $\mu=\E[x]$. Then
\begin{equation}
    z = W(x-\mu) + (b+W\mu).
    \label{eq:centered-affine}
\end{equation}
The matrix $W$ therefore serves two functions at once. It transforms deviations $x-\mu$, which determine sample-dependent shape, and it contributes $W\mu$, which is a sample-independent translation indistinguishable from a bias. The same overlap appears in the gradient and in the parameter update. A change $\Delta W$ contributes the shared displacement $\Delta W\mu$ even if the explicit bias is held fixed.

This observation is elementary algebra, and activation centering has a long history in conditioning and natural-gradient methods \citep{amari1998natural,desjardins2015natural,martens2015kfac}. What is not standard in modern optimizer analysis is to treat $\Delta W\mu$ as an explicit, measurable boundary channel and compare it directly with $\Delta b$ throughout training. Parameter-space summaries such as $\norm{\Delta W}$ and $\norm{\Delta b}$ cannot reveal which path actually moves the shared affine boundary.

This paper performs that functional decomposition in an instrumented Transformer. We make four contributions:
\begin{enumerate}
    \item We define a batch-level shape--boundary decomposition of affine gradients and updates. The centered shape displacement and the shared boundary displacement are exactly orthogonal in the empirical output-space inner product.
    \item We measure the hidden boundary channel in AdamW. Across 17 affine modules and 301 logged checkpoints, $\Delta W\mu$ is typically more than two orders of magnitude larger than the explicit bias step and explains almost all actual boundary displacement.
    \item We implement SBO-AdamW, a diagnostic optimizer that removes the bias-like component from the weight optimizer and assigns the boundary coordinate an independent Adam state. The prototype substantially accelerates fitting and improves the validation-selected test result in the present experiment.
    \item We identify a critical failure mode. Enforcing the decomposition in the original $(W,b)$ coordinates with a moving minibatch center causes large compensating bias updates and parameter gauge drift. This limitation separates the mechanism result from the design of a stable final algorithm.
\end{enumerate}

The evidence is intentionally presented as a focused mechanism study. It contains one dataset, one architecture, and one seed per method. The numerical improvements are therefore preliminary; the central result is the scale and persistence of hidden boundary motion under a standard optimizer.

\section{Background and Related Work}
\subsection{Adaptive optimization and tensor-wise geometry}
Adam normalizes a first-moment estimate by a diagonal second-moment estimate \citep{kingma2014adam}; AdamW decouples weight decay from this adaptive update \citep{loshchilov2017adamw}. Such methods operate coordinate-wise and do not explicitly model correlations between weight and bias. Muon applies matrix orthogonalization to hidden weight updates and has recently been scaled to large language-model training \citep{liu2025muon}. In common usage, biases, gains, embeddings, and heads remain under AdamW. This arrangement recognizes matrix geometry inside $W$ but does not remove the shared translation $\Delta W\mu$ that overlaps with the bias function.

\subsection{Natural gradients, centering, and whitening}
Natural gradient descent changes the parameter-space metric to reflect model geometry \citep{amari1998natural}. K-FAC approximates layerwise Fisher blocks by Kronecker factors and thereby retains non-diagonal layer structure \citep{martens2015kfac}. Natural Neural Networks and PRONG use function-preserving reparameterizations that center and whiten layer representations to improve Fisher conditioning \citep{desjardins2015natural}. These works establish that activation statistics and affine parameterization strongly affect optimization. Our decomposition is closely related algebraically, but our emphasis differs: we directly attribute the shared functional displacement to the weight and bias paths and track their responsibility over time.

\subsection{Gradient and weight constraints}
Gradient Centralization projects weight gradients onto zero-mean subspaces and interprets the operation as projected gradient descent \citep{yong2020gc}. Linearly Constrained Weights reduce activation shift caused by alignment between a weight vector and the mean activation vector \citep{kutsuna2024lcw}. Both are relevant to the present mechanism. However, the projection studied here is data- and bias-gradient-dependent:
\begin{equation}
    g_W^{\mathrm{shape}}=g_W-g_b\mu^\top,
\end{equation}
and its purpose is to remove the component that shares the exact boundary function of the explicit bias.

\subsection{Weight--bias responsibility}
Recent work on affine divergence analyzes mismatches between parameter updates and activation updates, and explicitly discusses the relative responsibility of weight and bias in propagated affine corrections \citep{bird2026affine}. Our study is complementary. We do not derive an ideal activation steepest-descent map; instead, we define the observable boundary channel $\Delta W\mu$, quantify it in a Transformer, and test a direct functional reassignment of that channel.

\section{Shape--Boundary Decomposition}
\subsection{Forward decomposition}
Let $x_i\in\R^{d_{\mathrm{in}}}$ be the input to an affine layer and let $W\in\R^{d_{\mathrm{out}}\times d_{\mathrm{in}}}$ and $b\in\R^{d_{\mathrm{out}}}$. For a minibatch of $n$ valid observations, define
\begin{equation}
    \mu = \frac{1}{n}\sum_{i=1}^n x_i,
    \qquad
    c = b+W\mu.
\end{equation}
Then
\begin{equation}
    z_i=W(x_i-\mu)+c.
    \label{eq:shape-boundary-forward}
\end{equation}
We call $W(x_i-\mu)$ the \emph{shape term}: its empirical mean over the observations is zero. We call $c$ the \emph{boundary coordinate}: it is shared by all observations.

For arbitrary changes $\Delta W$ and $\Delta c$, define
\begin{equation}
    \Delta z_i^{\mathrm{shape}}=\Delta W(x_i-\mu),
    \qquad
    \Delta z_i^{\mathrm{boundary}}=\Delta c.
\end{equation}
Their empirical output-space inner product is
\begin{align}
    \frac{1}{n}\sum_{i=1}^n
    \left\langle \Delta z_i^{\mathrm{shape}},
    \Delta z_i^{\mathrm{boundary}}\right\rangle
    &=
    \left\langle
    \Delta W\left(\frac{1}{n}\sum_{i=1}^n(x_i-\mu)\right),
    \Delta c
    \right\rangle \\
    &=0.
    \label{eq:function-orthogonality}
\end{align}
Thus, centered shape motion and shared boundary motion are exactly orthogonal for the observations used to define $\mu$.

\subsection{Gradient decomposition}
Let $\delta_i=\partial\mathcal{L}/\partial z_i$. With mean-reduced gradients,
\begin{equation}
    g_W = \frac{1}{n}\sum_i \delta_i x_i^\top,
    \qquad
    g_b = \frac{1}{n}\sum_i \delta_i.
\end{equation}
Substituting $x_i=(x_i-\mu)+\mu$ gives
\begin{equation}
    g_W =
    \underbrace{\frac{1}{n}\sum_i\delta_i(x_i-\mu)^\top}_{g_W^{\mathrm{shape}}}
    +
    \underbrace{g_b\mu^\top}_{g_W^{\mathrm{bias-like}}}.
    \label{eq:gradient-decomposition}
\end{equation}
Hence
\begin{equation}
    g_W^{\mathrm{shape}}=g_W-g_b\mu^\top.
\end{equation}
The second term is a rank-one component constructed from the bias gradient and the input mean. It is not necessarily orthogonal to $g_W^{\mathrm{shape}}$ in parameter space; the exact orthogonality applies to their induced shape and boundary displacements in output space.

\subsection{Update decomposition and diagnostics}
For an optimizer-generated parameter step $(\Delta W,\Delta b)$, the actual boundary displacement is
\begin{equation}
    \Delta c_{\mathrm{actual}}=\Delta b+\Delta W\mu.
    \label{eq:actual-boundary}
\end{equation}
We record the following quantities:
\begin{align}
    R_{\mathrm{grad}} &:= \frac{\norm{g_b\mu^\top}}{\norm{g_W}}, \\
    R_{\mathrm{leak,b}} &:= \frac{\norm{\Delta W\mu}}{\norm{\Delta b}}, \\
    R_{\mathrm{leak,c}} &:= \frac{\norm{\Delta W\mu}}{\norm{\Delta b+\Delta W\mu}}, \\
    G_{\mathrm{boundary}} &:= \frac{\norm{\Delta b+\Delta W\mu}}{\norm{\Delta b}}.
\end{align}
We additionally record the cosine between $\Delta b$ and $\Delta W\mu$, the shape and boundary function energies
\begin{equation}
    E_{\mathrm{shape}}
    =\frac{1}{n}\sum_i\norm{\Delta W(x_i-\mu)}_2^2,
    \qquad
    E_{\mathrm{boundary}}=\norm{\Delta c_{\mathrm{actual}}}_2^2,
\end{equation}
and the relative reconstruction and mapping errors used as numerical checks.

\section{Diagnostic SBO-AdamW}
The baseline applies ordinary AdamW directions to $g_W$ and $g_b$. Weight decay is set to zero in the present experiment, so the baseline is numerically equivalent to Adam with the AdamW implementation order.

SBO-AdamW instead maintains independent Adam states for the shape gradient and boundary gradient:
\begin{align}
    u_W &= \operatorname{AdamDirection}(g_W-g_b\mu^\top), \\
    u_c &= \operatorname{AdamDirection}(g_b), \\
    \Delta W &= -\eta u_W, \\
    \Delta c_{\mathrm{intended}} &= -\eta u_c.
\end{align}
The current prototype stores the original bias parameter and therefore applies the compensating step
\begin{equation}
    \Delta b=\Delta c_{\mathrm{intended}}-\Delta W\mu.
    \label{eq:sbo-compensation}
\end{equation}
This guarantees
\begin{equation}
    \Delta b+\Delta W\mu=\Delta c_{\mathrm{intended}}
\end{equation}
for the current minibatch center. This implementation is useful diagnostically because it isolates the functional channel while preserving a conventional affine forward pass. It is not the stable parameterization we ultimately recommend: because $\mu$ changes across minibatches, Eq.~\eqref{eq:sbo-compensation} can accumulate large canceling values in $b$.

A stable successor would store $(W,c)$ directly and evaluate
\begin{equation}
    z=W(x-\bar\mu)+c
\end{equation}
with a fixed or function-preservingly updated reference center $\bar\mu$. That design is discussed in Section~\ref{sec:discussion} but is not evaluated here.

\section{Experimental Setup}
\paragraph{Dataset.}
We use the IMDb binary sentiment dataset \citep{maas2011imdb}. The 25,000 labeled training reviews are split deterministically into 22,500 training and 2,500 validation examples; the standard 25,000-review test split is used only for reporting. A lowercase regular-expression tokenizer builds a vocabulary of at most 100,000 tokens with minimum frequency two. Sequences are truncated to length 256.

\paragraph{Model.}
The model is a four-layer pre-norm Transformer encoder \citep{vaswani2017attention,ba2016layernorm} with hidden dimension 256, four attention heads, feed-forward dimension 1024, dropout 0.1, sinusoidal positional encoding, and a two-class linear head. To obtain inputs for every affine map, attention projections are implemented explicitly rather than through a fused library module. The instrumented set contains 17 affine weight--bias pairs: QKV, attention output, feed-forward input, and feed-forward output in each of four layers, plus the classifier. These pairs contain 3,146,240 weight coordinates and 9,218 bias coordinates. Embeddings and normalization parameters remain under ordinary AdamW in both methods.

\paragraph{Optimization.}
Both runs use seed 42, batch size 128, Adam coefficients $(0.9,0.999)$, learning rate $10^{-3}$, epsilon $10^{-8}$, zero weight decay, gradient clipping at norm 1, and 3,000 optimizer steps. Full train, validation, and test evaluations are performed every 100 steps. The checkpoint is selected only by validation accuracy. All mechanism statistics are logged at step 1 and every 10 steps. No hyperparameter was tuned separately for SBO-AdamW.

\paragraph{Scope of evidence.}
This is a single-seed, single-task comparison. We report exact observations rather than means and confidence intervals. Any claim about general optimization behavior requires replication across seeds, architectures, datasets, and scale.

\section{Results}
\subsection{Predictive performance and training speed}
Table~\ref{tab:main-results} summarizes the validation-selected results. SBO-AdamW improves best validation accuracy by 4.13 percentage points and test accuracy at that checkpoint by 4.00 points. Its best validation checkpoint occurs at step 800, whereas AdamW continues improving until the final step.

\begin{table}[t]
\centering
\caption{Single-seed IMDb results. Test accuracy is reported at the checkpoint selected by validation accuracy.}
\label{tab:main-results}
\begin{tabular}{lrrrrrr}
\toprule
Method & Best step & Best val. & Train@best & Test@best & Final val. & Final test \\
\midrule
AdamW & 3000 & 81.68 & 87.93 & 78.73 & 81.68 & 78.73 \\
SBO-AdamW & 800 & \textbf{85.81} & \textbf{94.62} & \textbf{82.73} & \textbf{82.92} & \textbf{80.72} \\
\bottomrule
\end{tabular}
\end{table}

\begin{figure}[t]
\centering
\includegraphics[width=0.78\linewidth]{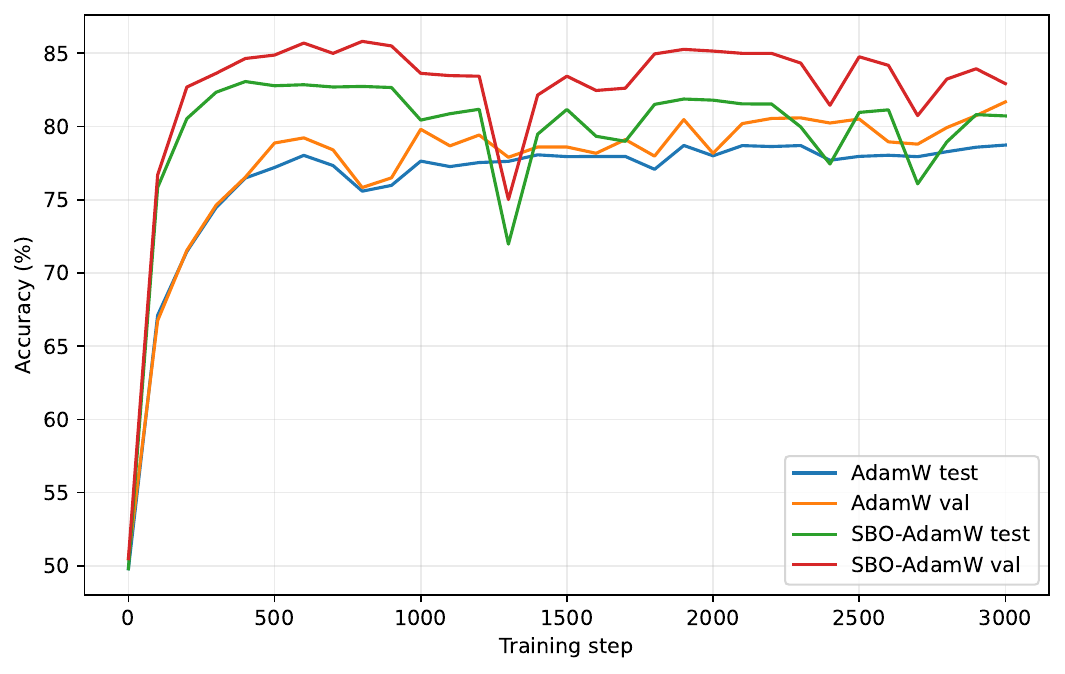}
\caption{Validation and test accuracy over training. SBO-AdamW fits substantially faster but begins to overfit after its validation optimum. Curves are from one seed.}
\label{fig:accuracy}
\end{figure}

The early difference is large. At step 200, the test accuracies are 71.46\% for AdamW and 80.53\% for SBO-AdamW. At step 400 they are 76.48\% and 83.07\%, respectively. The later SBO trajectory shows a widening train--test gap: train accuracy reaches 98.52\% while final test accuracy falls to 80.72\%. Thus, the prototype accelerates useful representation learning but also accelerates late-stage overfitting.

\subsection{The raw weight gradient contains a large bias-like component}
Across all AdamW affine-layer checkpoints, the median gradient-overlap ratio is
\begin{equation}
    \operatorname{median}(R_{\mathrm{grad}})= 0.664,
\end{equation}
with a 10th--90th percentile range of [0.319, 0.944]. The median cosine between $g_W$ and $g_b\mu^\top$ is 0.556. The components are therefore not merely both large; they are often positively aligned.

\begin{figure}[t]
\centering
\includegraphics[width=0.88\linewidth]{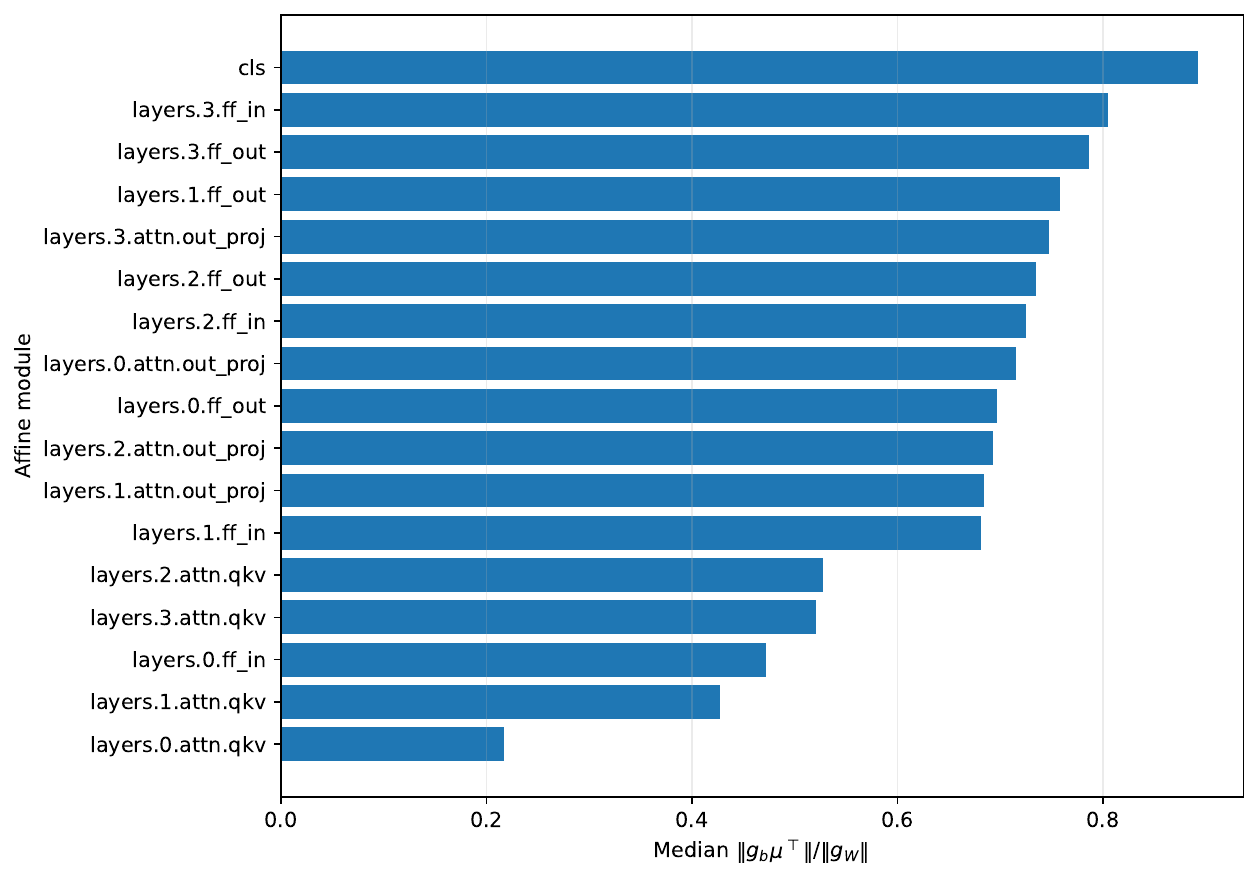}
\caption{Median bias-like gradient magnitude by affine module under AdamW. The ratio is large in most modules and especially high in the classifier and deeper feed-forward/output projections.}
\label{fig:gradient-overlap}
\end{figure}

This ratio is not an energy partition because $g_W^{\mathrm{shape}}$ and $g_b\mu^\top$ need not be orthogonal in parameter space. Its significance is functional: the raw matrix gradient contains a substantial component generated by the same batch-mean signal as the bias gradient.

\subsection{AdamW boundary motion is dominated by the weight path}
The update decomposition is more extreme than the gradient decomposition. Across affine-layer checkpoints under AdamW,
\begin{equation}
    \operatorname{median}(R_{\mathrm{leak,b}})= 134.7,
\end{equation}
with a 10th--90th percentile range of [45.0, 740.0]. The median ratio to the actual boundary displacement is
\begin{equation}
    \operatorname{median}(R_{\mathrm{leak,c}})= 0.9944.
\end{equation}
The median cosine between the explicit bias step and the weight-induced boundary step is 0.892, so the two paths usually reinforce rather than cancel each other. The corresponding median boundary interference gain is 135.6.

\begin{figure}[t]
\centering
\includegraphics[width=0.9\linewidth]{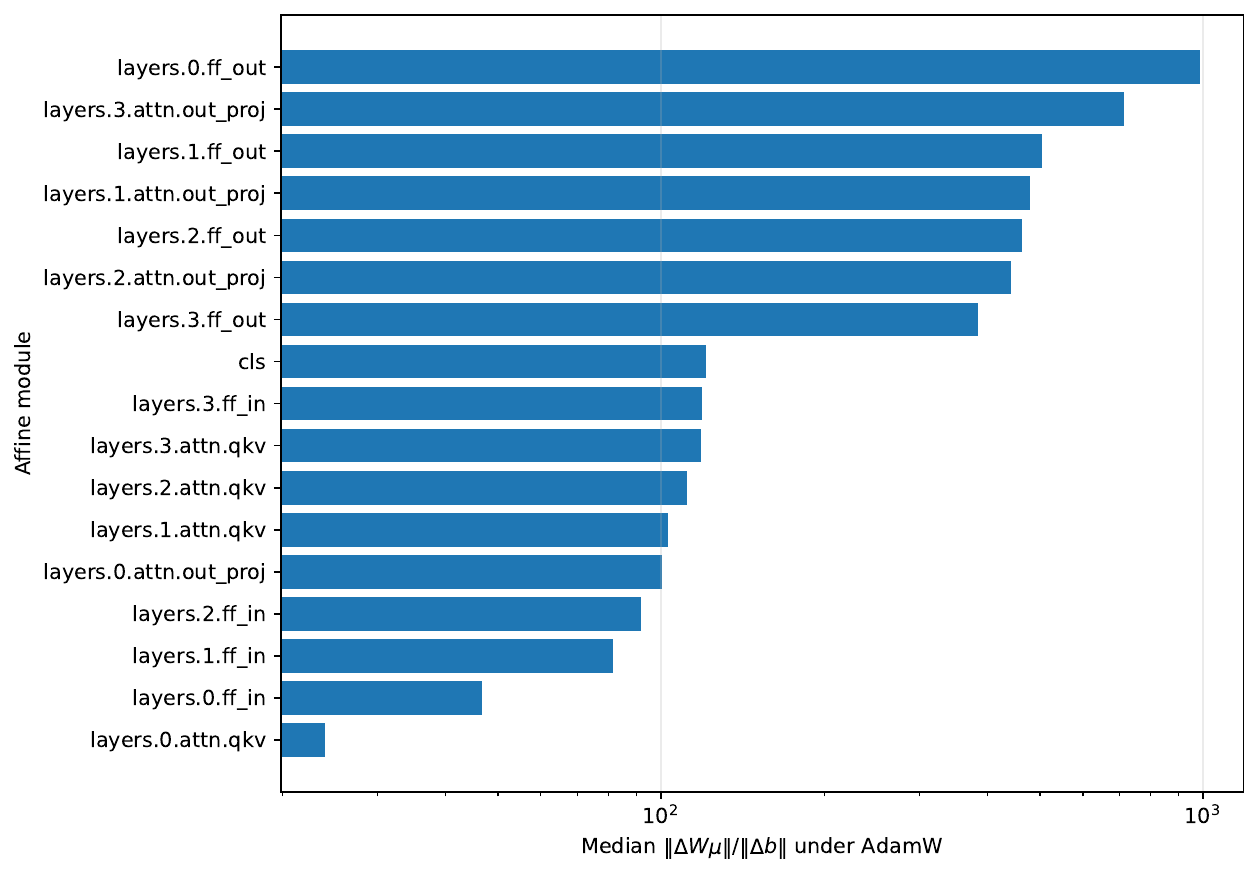}
\caption{Median hidden boundary leakage under AdamW. Every affine module has $\norm{\Delta W\mu}$ far larger than the explicit bias step; output projections and feed-forward output layers are most affected. The horizontal axis is logarithmic.}
\label{fig:layer-leakage}
\end{figure}

\begin{figure}[t]
\centering
\includegraphics[width=0.78\linewidth]{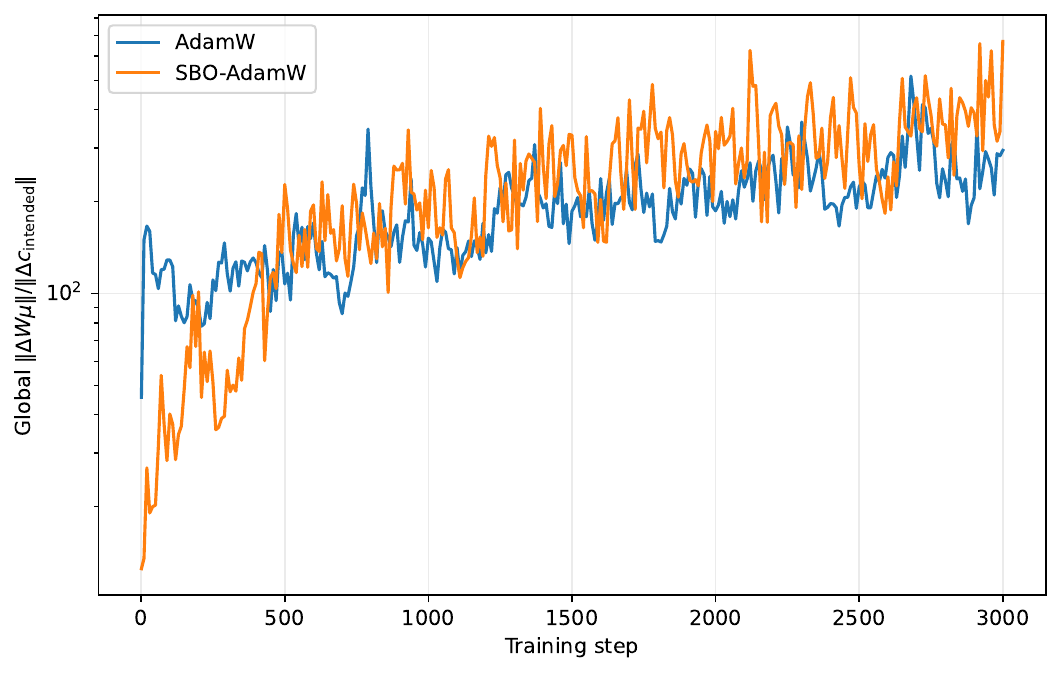}
\caption{Global hidden-boundary ratio over training. In the SBO run, the ratio measures the size of the compensation that must be canceled to realize the independently optimized boundary update; it is not residual functional leakage.}
\label{fig:global-leakage}
\end{figure}

At the final AdamW step, the global ratio $\norm{\Delta W\mu}/\norm{\Delta b}$ is 294.5. These observations support a responsibility inversion: although $b$ is the named translation parameter, the shared output displacement is realized primarily by $W$.

\subsection{SBO enforces functional reassignment}
The gradient decomposition reconstructs the raw weight gradient to numerical precision in both runs: the median relative reconstruction error is $2.31e-08$ under SBO-AdamW. More importantly, Eq.~\eqref{eq:sbo-compensation} is realized with a median boundary mapping error of $2.36e-06$. The median boundary interference gain is exactly 1.000000, and the median cosine between the stored bias step and $\Delta W\mu$ is -0.999982. Thus, the large weight-induced boundary displacement is present in the original coordinates but is almost exactly canceled, leaving the independent $\Delta c$ as the realized functional boundary update.

The centered shape and boundary function displacements have near-zero measured cosine in both runs, as required by Eq.~\eqref{eq:function-orthogonality}. This fact alone does not distinguish the methods. The distinction is that ordinary AdamW uses the same $\Delta W$ to produce both the centered shape term and the hidden boundary term, whereas SBO-AdamW explicitly removes the latter from the realized function update.

\subsection{Boundary suppression is a confound, not a negligible detail}
SBO-AdamW changes more than attribution. It also changes the scale of the realized boundary channel. At step 800, the summed boundary function energy is 17.9 under AdamW but only $5.21e-05$ under SBO-AdamW. Meanwhile the shape energies are 8.47 and 2.73. Therefore, the current performance difference cannot be attributed purely to orthogonalization: the prototype effectively removes the large AdamW boundary amplifier and leaves a much smaller independently optimized boundary step.

\begin{figure}[t]
\centering
\includegraphics[width=0.78\linewidth]{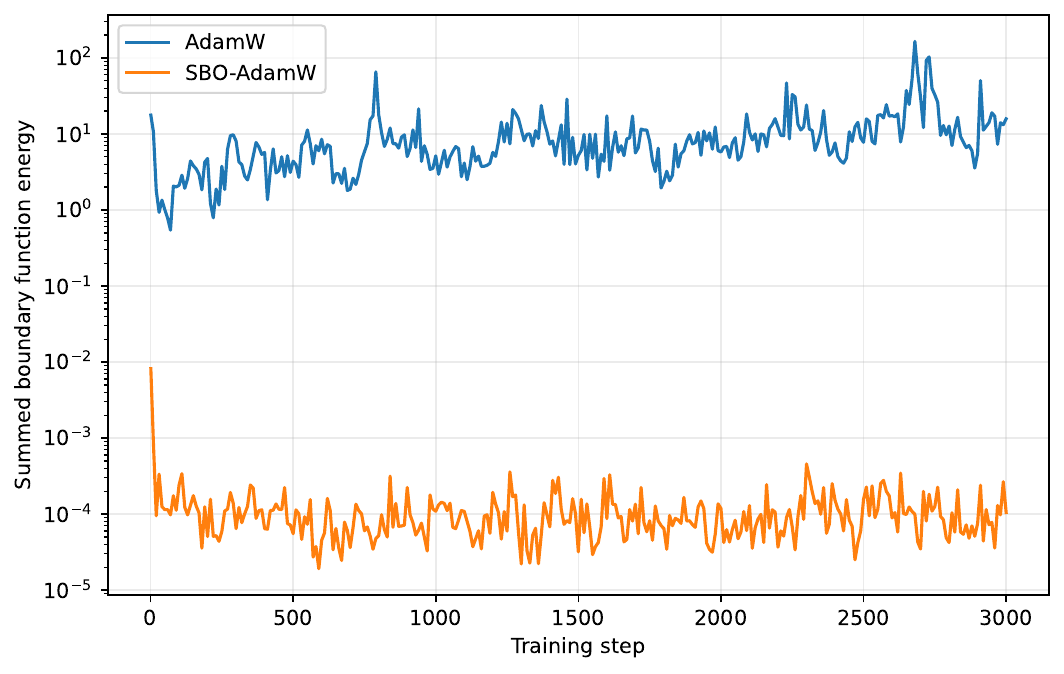}
\caption{Summed boundary function energy. The current SBO prototype strongly suppresses realized boundary motion relative to AdamW. This is a central confound that future amplitude-matched and stable-parameterization experiments must isolate.}
\label{fig:boundary-energy}
\end{figure}

This result is nevertheless mechanistically informative. If the hidden $\Delta W\mu$ path were irrelevant, removing it would not be expected to alter optimization so strongly. The experiment shows that its removal changes both the speed and generalization trajectory, while leaving open whether the beneficial cause is cleaner functional separation, reduced boundary amplitude, changed Adam moments, or a combination of these factors.

\subsection{Moving-center compensation causes parameter gauge drift}
The current implementation represents the function in the original bias coordinate while changing the center every minibatch. When $\Delta W\mu$ is large, Eq.~\eqref{eq:sbo-compensation} requires a comparably large opposite change in $b$ even though the realized boundary step is small. The median affine bias norm over logged SBO layer-steps is 21.75, compared with 0.766 under AdamW. The maximum observed SBO bias norm is 673.63, versus 1.96 for AdamW.

\begin{figure}[t]
\centering
\includegraphics[width=0.78\linewidth]{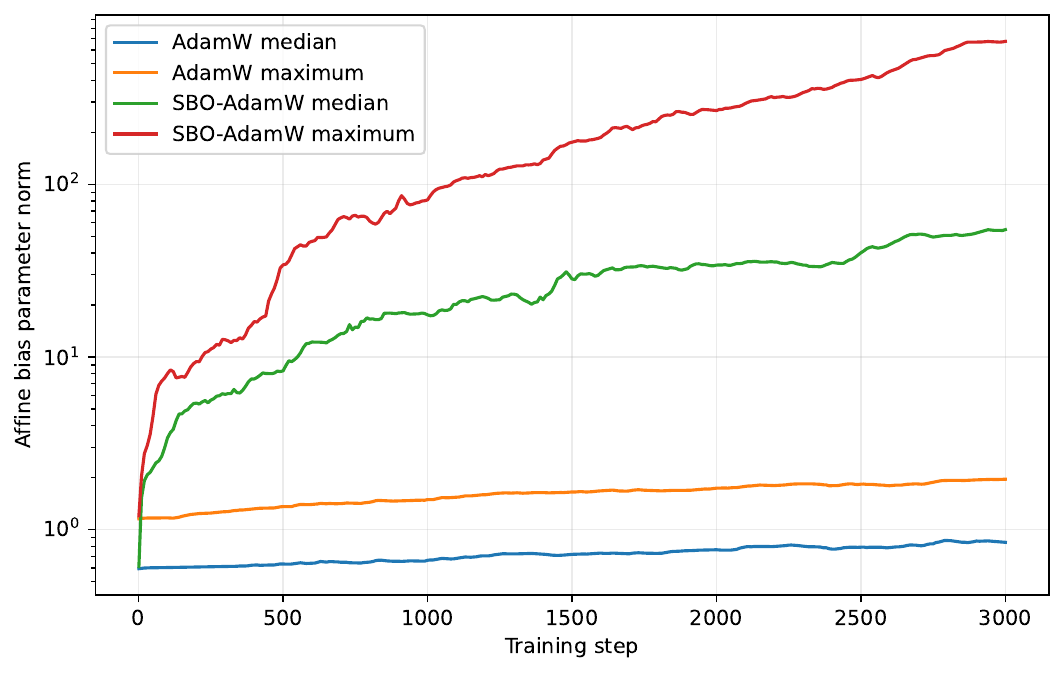}
\caption{Affine bias parameter norms. The compensation-based SBO implementation develops severe gauge drift, even though the realized boundary update remains small. The vertical axis is logarithmic.}
\label{fig:bias-growth}
\end{figure}

This is not an unavoidable cost of function-space orthogonalization. It is a consequence of retaining $b$ while repeatedly redefining $c=b+W\mu$ with a moving center. A stable implementation should store $c$ directly.

\section{Discussion}
\label{sec:discussion}
\subsection{Parameter type is not functional type}
The central empirical observation is that a matrix parameter can contain a vector-like functional channel. Under a nonzero input mean, $\Delta W\mu$ is shared across all observations and is therefore functionally a boundary translation. Reporting only weight and bias update norms obscures this channel and can lead to the misleading conclusion that biases are unimportant because $\norm{\Delta b}$ is small. In the present AdamW run, $\norm{\Delta b}$ is small precisely because the weight path already supplies a much larger aligned translation.

This distinction also changes how optimizer specialization should be interpreted. A matrix optimizer can improve the geometry internal to $W$ while still allowing the matrix update to dominate the boundary. Conversely, assigning $b$ to AdamW does not give the explicit bias control over the function it nominally represents. A functional optimizer should first identify the shape and boundary subspaces, and only then choose geometry within each subspace.

\subsection{Relationship to centering and natural-gradient methods}
Equation~\eqref{eq:centered-affine} is not a new algebraic identity, and centered or whitened reparameterizations have been studied for conditioning and natural-gradient approximation \citep{desjardins2015natural,martens2015kfac}. The present contribution is narrower and empirical: it treats $\Delta W\mu$ as a first-class training signal, quantifies its responsibility relative to $\Delta b$, and shows that changing this responsibility produces a large optimization effect in a Transformer.

The results also complement activation-shift constraints \citep{kutsuna2024lcw} and affine-divergence analysis \citep{bird2026affine}. Those lines of work motivate controlling affine geometry; our diagnostics identify a concrete hidden channel through which standard adaptive updates move shared boundaries.

\subsection{Toward a stable centered-affine optimizer}
A stable version should parameterize each affine layer as
\begin{equation}
    z=W(x-\bar\mu)+c,
    \label{eq:stable-centered}
\end{equation}
where $c$ is stored directly. The reference center $\bar\mu$ can be estimated during a calibration period and frozen, or updated occasionally with a function-preserving transformation. If the center changes from $\bar\mu_{\mathrm{old}}$ to $\bar\mu_{\mathrm{new}}$, preserving the affine function requires
\begin{equation}
    c_{\mathrm{new}} = c_{\mathrm{old}} + W(\bar\mu_{\mathrm{new}}-\bar\mu_{\mathrm{old}}).
\end{equation}
Autodifferentiation through Eq.~\eqref{eq:stable-centered} then produces the centered shape gradient directly, without post-hoc subtraction or large canceling bias updates.

Once the parameterization is stable, different optimization geometries become meaningful. A matrix method such as Muon can be applied to the shape coordinate $W$, while Adam can optimize the boundary vector $c$. This would combine weight-internal matrix geometry with weight--bias functional separation. The present experiment does not test that design.

\subsection{Required ablations}
Four experiments are necessary before claiming a general optimizer:
\begin{enumerate}
    \item \textbf{Stable parameterization:} compare stored-$(W,c)$ centered affine layers with the compensation prototype.
    \item \textbf{Boundary-amplitude matching:} match $E_{\mathrm{boundary}}$ across methods to separate functional reassignment from amplitude suppression.
    \item \textbf{Optimizer-state ablation:} compare shared and independent moment states for the decomposed channels.
    \item \textbf{Cross-domain validation:} repeat across MLPs, CNNs, Transformers, multiple datasets, multiple seeds, and pretrained as well as from-scratch settings.
\end{enumerate}

\section{Limitations}
The study has several substantial limitations.
\begin{itemize}
    \item Only one seed, one dataset, and one small Transformer are evaluated. No confidence interval or significance test is possible.
    \item The current SBO implementation uses the current minibatch mean. Its exact orthogonality is local to that batch and its coordinate system moves over training.
    \item The compensation implementation causes severe bias-coordinate growth. This can affect numerical conditioning and late-stage optimization.
    \item SBO changes the realized boundary amplitude by orders of magnitude. The performance gain is not a clean causal estimate of orthogonality alone.
    \item The explicit attention implementation was required for instrumentation. Although both methods use the same architecture, results should not be compared directly with unrelated stock-Transformer runs.
    \item The study does not yet compare against PRONG, K-FAC, Gradient Centralization, linearly constrained weights, Muon, or affine-divergence corrections in the same experimental protocol.
\end{itemize}
These limitations are not secondary details. They define the next experimental stage and constrain the claims of the present manuscript to mechanism discovery and prototype validation.

\section{Conclusion}
An affine weight matrix does not exclusively learn sample-dependent shape. When its input has nonzero mean, every weight update also produces a shared boundary displacement $\Delta W\mu$. In the instrumented AdamW Transformer studied here, this hidden channel is much larger than the explicit bias step and accounts for nearly all realized boundary motion. A diagnostic optimizer that removes the bias-like weight-gradient component and independently controls the boundary changes the training trajectory dramatically, improving the validation-selected test accuracy by about four percentage points and reaching its best validation checkpoint much earlier.

The same prototype exposes the unresolved issue: moving-center compensation suppresses boundary energy and creates large parameter drift. The next method should therefore not be another scalar amplitude controller. It should use a stable centered-affine parameterization that stores shape and boundary coordinates separately, then assign an appropriate geometry to each. The broader hypothesis is that functional roles, not tensor shapes alone, should determine how neural-network parameters are optimized.

\appendix
\section{Proof of Empirical Function-Space Orthogonality}
Let $X_c=[x_1-\mu,\ldots,x_n-\mu]$ denote centered inputs and let the boundary displacement be constant across observations. Using the Frobenius inner product over the batch,
\begin{align}
\left\langle \Delta Z_{\mathrm{shape}},\Delta Z_{\mathrm{boundary}}\right\rangle_F
&=\sum_{i=1}^n
\left\langle \Delta W(x_i-\mu),\Delta c\right\rangle\\
&=\left\langle \Delta W\sum_{i=1}^n(x_i-\mu),\Delta c\right\rangle\\
&=0.
\end{align}
No condition on the optimizer or the relative magnitudes of the two updates is required. The statement is exact for the observations used to compute $\mu$.

\section{Complete AdamW Layer Summary}
Table~\ref{tab:all-layers} reports medians over all logged checkpoints. The leakage ratio is large in every affine module, including the module with the smallest value.

\begin{table}[h]
\centering
\small
\caption{AdamW per-module medians across 301 logged checkpoints. Modules are ordered by hidden boundary leakage.}
\label{tab:all-layers}
\begin{tabular}{r>{\raggedright\arraybackslash}p{0.32\linewidth}rrrr}
\toprule
Rank & Module & $R_{\mathrm{grad}}$ & Grad. cosine & $R_{\mathrm{leak,b}}$ & $R_{\mathrm{leak,c}}$ \\
\midrule
1 & \texttt{layers.0.ff\_out} & 0.697 & 0.423 & 985.4 & 0.9992 \\
2 & \texttt{layers.3.attn.out\_proj} & 0.747 & 0.804 & 715.5 & 0.9988 \\
3 & \texttt{layers.1.ff\_out} & 0.758 & 0.657 & 505.3 & 0.9983 \\
4 & \texttt{layers.1.attn.out\_proj} & 0.684 & 0.580 & 479.1 & 0.9985 \\
5 & \texttt{layers.2.ff\_out} & 0.734 & 0.740 & 463.3 & 0.9981 \\
6 & \texttt{layers.2.attn.out\_proj} & 0.692 & 0.648 & 442.4 & 0.9983 \\
7 & \texttt{layers.3.ff\_out} & 0.786 & 0.882 & 385.0 & 0.9975 \\
8 & \texttt{cls} & 0.892 & 0.941 & 121.1 & 0.9919 \\
9 & \texttt{layers.3.ff\_in} & 0.804 & 0.876 & 119.0 & 0.9920 \\
10 & \texttt{layers.3.attn.qkv} & 0.520 & 0.470 & 118.7 & 0.9925 \\
11 & \texttt{layers.2.attn.qkv} & 0.527 & 0.447 & 111.6 & 0.9924 \\
12 & \texttt{layers.1.attn.qkv} & 0.427 & 0.309 & 102.8 & 0.9928 \\
13 & \texttt{layers.0.attn.out\_proj} & 0.715 & 0.238 & 100.5 & 0.9943 \\
14 & \texttt{layers.2.ff\_in} & 0.725 & 0.683 & 91.9 & 0.9905 \\
15 & \texttt{layers.1.ff\_in} & 0.681 & 0.568 & 81.7 & 0.9907 \\
16 & \texttt{layers.0.ff\_in} & 0.472 & 0.214 & 46.6 & 0.9894 \\
17 & \texttt{layers.0.attn.qkv} & 0.217 & 0.191 & 24.0 & 0.9632 \\
\bottomrule
\end{tabular}
\end{table}

\section{Selected Checkpoint Values}
\begin{table}[h]
\centering
\caption{Accuracy and function-energy snapshots. These values are descriptive and come from one seed.}
\begin{tabular}{lrrrrrr}
\toprule
Method & Step & Train acc. & Val. acc. & Test acc. & $E_{\mathrm{shape}}$ & $E_{\mathrm{boundary}}$ \\
\midrule
AdamW & 200 & 72.75 & 71.54 & 71.46 & 1.74 & 4.76 \\
SBO-AdamW & 200 & 84.66 & 82.69 & 80.53 & 2.29 & 5.07e-05 \\
AdamW & 800 & 77.74 & 75.83 & 75.58 & 8.47 & 17.9 \\
SBO-AdamW & 800 & 94.62 & 85.81 & 82.73 & 2.73 & 5.21e-05 \\
AdamW & 3000 & 87.93 & 81.68 & 78.73 & 6.46 & 15.8 \\
SBO-AdamW & 3000 & 98.52 & 82.92 & 80.72 & 122 & 0.000105 \\
\bottomrule
\end{tabular}
\end{table}

\section{Reproducibility Notes}
The supplementary directory contains the raw extracted CSV/JSON logs, 13 plot-ready CSV files, generated figures, and a plotting script. The baseline optimizer was numerically checked against PyTorch AdamW under identical gradients in the implementation workflow. The SBO reconstruction and boundary-mapping metrics are logged for every affine layer and checkpoint. The exact training code is provided separately as \texttt{step21\_train\_imdb\_shape\_boundary\_orthogonal\_adam.py}.

\end{document}